\documentclass[twoside,11pt]{article}

\usepackage{./jmlr2e}

\usepackage{listings}
\usepackage{graphicx}
\usepackage{subcaption}
\graphicspath{ {./images/} }
\usepackage{natbib}

\ShortHeadings{SMART}{Chew, Wenger, Kery, Nance, Richards, Hadley, and Baumgartner}
\firstpageno{1}

\begin{document}
\raggedbottom

\title{SMART: An Open Source Data Labeling Platform for Supervised Learning}

\author{\name Rob Chew \email rchew@rti.org\\ 
\name Michael Wenger \email mwenger@rti.org\\ 
\name Caroline Kery \email ckery1@umbc.edu\\ 
\name Jason Nance \email jnance@rti.org\\ 
\name Keith Richards \email krichards@rti.org\\ 
\name Emily Hadley \email ehadley@rti.org\\ 
\name Peter Baumgartner \email pbaumgartner@rti.org\\ 
\addr Center for Data Science\\ 
RTI International\\  
Research Triangle Park, NC 27709, USA}
\editor{TBD}

\maketitle

\begin{abstract}%
SMART is an open source web application designed to help data scientists and research teams efficiently build labeled training data sets for supervised machine learning tasks. SMART provides users with an intuitive interface for creating labeled data sets, supports active learning to help reduce the required amount of labeled data, and incorporates inter-rater reliability statistics to provide insight into label quality. SMART is designed to be platform agnostic and easily deployable to meet the needs of as many different research teams as possible. The project website \footnote{https://rtiinternational.github.io/SMART/} contains links to the code repository and extensive user documentation.
\end{abstract}


\section{Introduction}
Over the past decade, supervised machine learning has made tremendous strides, achieving or surpassing human-level performance on tasks diverse as image recognition \citep{real2017large}, speech recognition \citep{saon2017english} and language modeling \citep{zilly2016recurrent}. While many of these achievements are at least partially due to algorithmic advances and specialized computing resources, they also all rely on large labeled data sets. However, in the research community and in industry, it is often acknowledged that the main bottleneck in machine learning adoption is no longer algorithms or hardware, but creating sufficiently large labeled data sets to address meaningful tasks.

To address this issue, we created SMART \footnote{SMART stands for “Smarter Manual Annotation for Resource-constrained collection of Training data”, an acronym so contrived that it qualifies as both a backronym and a recursive acronym.}, an open source application designed to lessen manual labeling effort while maintaining data quality. SMART uses active learning \citep{settles2012active} to help reduce the number of labels necessary to develop a reliable model, inter-rater reliability statistics \citep{doi:10.1177/001316446002000104,fleiss1971measuring} to ensure label quality, and data visualization to help project administrators manage coding projects. SMART allows for multiple coders per project and supports on-premise installation, essential for projects where crowdsourcing is prohibited (e.g., when the data set is proprietary or sensitive in nature). While SMART was primarily designed for applied researchers and analysts needing to generate labelled training data, machine learning and human-computer interaction researchers can also use SMART to study how humans label data. This could range from running experiments on how coders interact to assessing active learning algorithms on their total labelling burden (time, costs, etc.).

SMART is part of a broad collection of commercial and open source data labeling platforms. While providing some overlap in functionality, these software products and services range widely in their intended use cases and pricing. Commercial Platform as a Service (PaaS) applications that specialize in data labelling (e.g., Figure Eight) tend to be the most full-featured and put a premium on the usability and scalability of their products. For a fee, these services provide analysts with a platform to organize, annotate, and export their labeled data. In contrast, commercial crowdsourcing platforms (e.g., Amazon Mechanical Turk) provide both annotation templates and access to an on-demand labelling workforce. While crowdsourcing may be an inexpensive and reliable way to perform many micro-tasks, it may be inappropriate for more complex tasks, those that require expert judgement, or for data that is confidential or classified. For open source projects, specialized research software applications focusing on methods like active learning \citep{settles2011closing} and adaptive sampling \citep{scott_sievert-proc-scipy-2017} are more common than platforms taking a holistic approach to the data labelling workflow. In essence, SMART is the open source alternative to commercial data labelling platforms, providing a powerful set of core features that can be modified and extended by the community to address new use cases. 

\section{Using SMART}
SMART is built using an open-source technology stack with the intention of being platform agnostic and easy to deploy. Specifically, SMART is a web application using React, Bootstrap, d3, and webpack for front-end interactions and Django, Redis, and PostgreSQL to support back-end operations. Docker and Docker-Compose \citep{docker} are used for OS-level virtualization to aid configuration management and ease deployment in new environments. Docker and Docker-Compose are also SMART’s only dependencies; instructions to install SMART using these systems can be found online in the user documentation \footnote{https://smart-app.readthedocs.io/en/latest/tutorial-installation.html}.

{\bf Project Creation.} Users create their own projects using a multi-step workflow. To start a project, users provide basic project information (project name and description), declare the set of classes they will annotate in the project, and upload their unlabeled data. Optionally, they can also set user permissions for multi-user projects, modify advanced project settings, and upload an instructional codebook to help define labeling tasks for users. When uploading project data, users have the option of providing a unique identifier to help link exported records and they can also provide pre-labeled data to initialize the active learning models. The advanced settings allow the user to customize the active learning model, set the batch size for the number of observations to label prior to re-running computations, and to enable/disable inter-rater reliability. While sensible default settings are provided, all settings can be customized to meet the specific needs of the labeling project. At the time of writing, SMART primarily supports text classification, though it can be extended to model other notable tasks and data types.

{\bf Annotating Data.} SMART’s annotation page allows users to perform several different tasks relating to labeling data. All project members have access to the main labeling interface (Figure 1a) as well as their personal annotation history where they can view and modify previously labeled records. For each record, users can either provide a label or skip the observation. Skipped records are forwarded to a project administrator for adjudication. To help with the labeling task, the annotation page also contains the instructional codebook and the project’s labels and definitions (if included during project creation), providing easy access to important annotation resources. Privileged users can also label specific records if the label distribution is highly skewed \citep*{Attenberg:2010:WLY:1835804.1835859}, resolve skipped data elements, and remove records from the project that are completely out-of-scope or otherwise deemed unusable. If inter-rater reliability is enabled, privileged users can also adjudicate double-coded records in which there is label disagreement.

\begin{figure}[ht] 
  \begin{subfigure}[b]{0.5\linewidth}
    \centering
    \includegraphics[width=0.95\linewidth,height=0.75\linewidth]{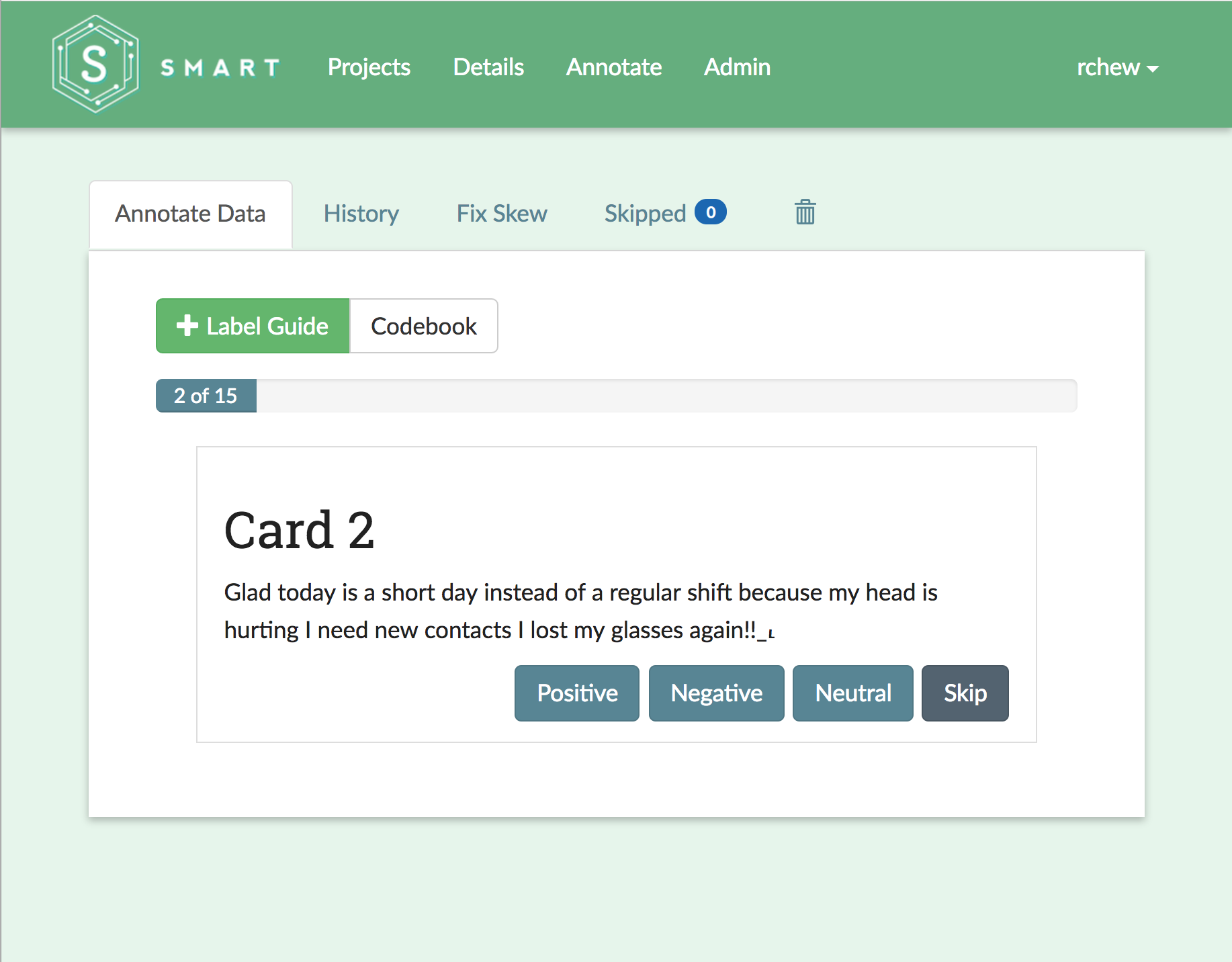} 
    \caption{Annotation Page} 
    \label{fig1:a} 
    \vspace{1ex}
  \end{subfigure}%% 
  \begin{subfigure}[b]{0.5\linewidth}
    \centering
    \includegraphics[width=0.95\linewidth,height=0.75\linewidth]{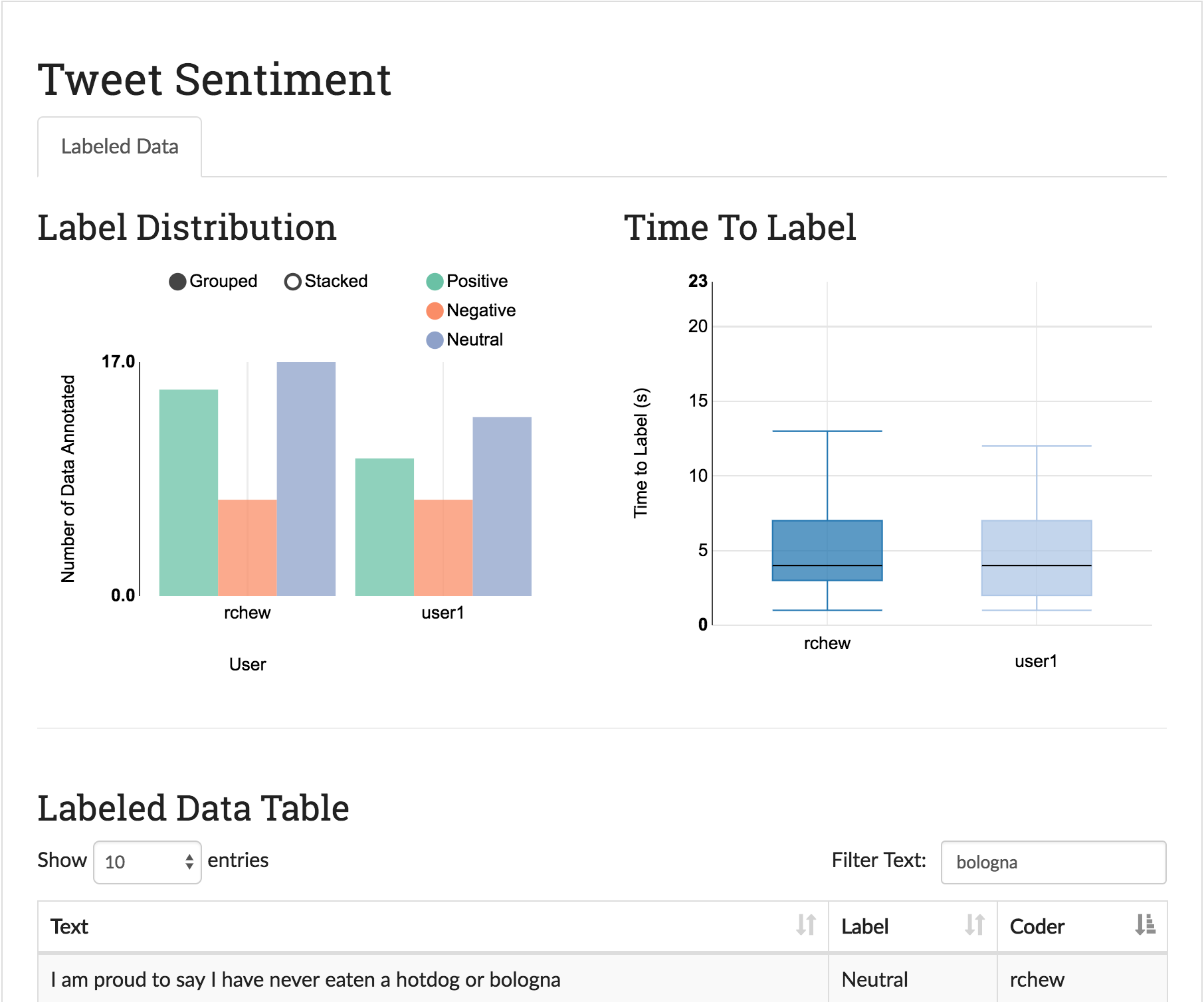} 
    \caption{Coder Visualization} 
    \label{fig1:b} 
    \vspace{1ex}
  \end{subfigure} 
  \caption{Screenshot of various application pages.}
  \label{fig1} 
\end{figure}

{\bf Administrative Dashboard.} To help project owners manage their labeling tasks, privileged users have access to an administrator dashboard page that includes data visualizations of the labeling process, performance metrics for the active learning model, and inter-rater reliability metrics across labelers.

The main labeled data view includes two interactive visualizations and an informative data table (Figure 1b). The label distribution chart shows the count of each label per user to compare labeler progress and identify drastic differences in label distributions that might indicate a user’s misunderstanding of the labeling task. The time to label chart is a box and whisker plot detailing how long each user is on the annotation page before selecting a label. This chart can help indicate if certain labelers are taking much less time than expected (indicating possible falsification) or more time than expected (indicating potential barriers to labeling).

The active learning model view provides information on the performance of a classifier trained on the labeled project data. The purpose of this view is to help provide insight on when the data labeling has hit a natural saturation point (additional labels are not improving model performance) and to add transparency to the active learning mechanism. SMART currently supports uncertainty sampling \citep{lewis1994sequential} as the active learning algorithm using least confident, margin, and entropy as measures of uncertainty (Settles, 2012). Currently, several classifier types are supported through the Scikit-learn python library \citep{pedregosa2011scikit}, and text data elements are vectorized using a term frequency-inverse document frequency (tf-idf) representation \citep{salton1988term}. After each batch of data is labeled, the model is retrained, and its performance is updated and displayed in an interactive data visualization.

Finally, the inter-rater reliability (IRR) view lets privileged users understand how consistently labelers agree on labels that are double-coded. This view is particularly useful early in the labeling process to discover if there is ambiguity in the class definitions or if the task is underspecified. It also provides an indication of how well humans would be expected to perform on the task, a useful benchmark to contextualize model performance. SMART uses Cohen’s kappa coefficient \citep{doi:10.1177/001316446002000104} to measure IRR between two independent labelers and Fleiss’ kappa \citep{fleiss1971measuring} for more than two labelers. Users can also view the percent overall agreement and the pairwise percent agreement for unadjusted metrics comparing consistency across labelers. As a visual summary, SMART provides a confusion matrix of labeler agreement in the form of a dynamic heat map. This visualization is valuable in diagnosing if there are particular classes in which labelers disagree, providing targeted insight into potentially ambiguous class definitions.

{\bf Exporting Data.} SMART allows users to download both the labeled data and the underlying classification model used for active learning (if enabled for the project). For all projects, SMART returns a zipped file of labeled data, sorted by label. If active learning is enabled, SMART will also provide several relevant serialized Python objects in Pickle format, including the preprocessed version of the input data as a tf-idf matrix, the trained classifier, and the fit vectorizer to convert text data into tf-idf vectors. Additionally, a README pdf is included that describes the files and contains sample Python code showing how to preprocess new data and use the trained model to predict on new observations.

\acks{
We would like to acknowledge the National Consortium for Data Science and RTI International for their generous support in funding the development of SMART. We would also like to thank Lucy Liu for creation of the SMART logo as well as valuable feedback, guidance, and user testing from Annice Kim, Maggie Cress, Jennifer Unangst, and Joey Morris.}

\bibliography{smart.bib}

\end{document}